\documentclass{article}

\PassOptionsToPackage{numbers, compress}{natbib}

\usepackage[final]{tackling_climate_workshop_style}

\usepackage[utf8]{inputenc} 
\usepackage[T1]{fontenc}    
\usepackage{hyperref}       
\usepackage{url}            
\usepackage{booktabs}       
\usepackage{amsfonts}       
\usepackage{nicefrac}       
\usepackage{microtype}      

\usepackage{amsmath}
\usepackage{amssymb}
\usepackage{xcolor}

\usepackage{cleveref}
\usepackage{graphicx}
\usepackage{multirow}
\usepackage{caption}
\usepackage{subcaption}
\usepackage{paralist} 
\usepackage{wrapfig}
\usepackage{mathrsfs} 
\usepackage{enumitem} 

\newcommand{\stateparnull}{\ensuremath{ {w} }} 

\newcommand{\statedis}[1]{\ensuremath{ {\stateparnull} }} %
\newcommand{\normL}[1]{\ensuremath{ \mathcal{L}_{#1}} }

\newcommand{\Rey}{\ensuremath{ \text{Re} }}

\usepackage[acronym]{glossaries}
\newacronym{medida}{MEDIDA}{Model Error Discovery with Interpretability and Data Assimilation}
\newacronym{da}{DA}{data assimilation}
\newacronym{enkf}{EnKF}{ensemble Kalman filter}
\newacronym{enks}{EnKS}{ensemble Kalman smoother}
\newacronym{pde}{PDE}{partial differential equation}
\newacronym{ode}{ODE}{ordinary differential equation}
\newacronym{dns}{DNS}{direct numerical simulation}
\newacronym{les}{LES}{large eddy simulation}
\newacronym{sgs}{SGS}{sub-grid scale}
\newacronym{eki}{EKI}{Ensemble Kalman Inversion}
\newacronym{rl}{RL}{reinforcement learning}
\newacronym{marl}{SMARL}{Scientific Multi-Agent Reinforcement Learning}

\newacronym{rvm}{RVM}{relevance vector machine}
\newacronym{lasso}{LASSO}{least absolute shrinkage and selection operator}
\newacronym{sindy}{SINDy}{sparse identification of nonlinear dynamics}
\newacronym{ml}{ML}{machine learning}
\newacronym{nn}{NN}{deep neural network}
\newacronym{mlp}{MLP}{multi-layer perceptron}
\newacronym{ntk}{NTK}{neural tangent kernel}
\newacronym{dnn}{DNN}{deep neural network}
\newacronym{ann}{ANN}{artificial neural network}
\newacronym{cnn}{CNN}{convolutional neural network}
\newacronym{ks}{KS}{Kuramoto-Sivashinsky}
\newacronym{etdrk4}{ETDRK4}{exponential time differencing fourth-order Runge-Kutta}
\newacronym{gcm}{GCM}{global climate model}
\newacronym{nwp}{NWP}{numerical weather prediction}
\newacronym{mooam}{MAOOAM}{modular arbitrary-order-ocean-atmosphere model}
\newacronym{qg}{QG}{quasi-geostrophic}
\newacronym{rff}{RFF}{random Fourier feature}
\newacronym{rmse}{RMSE}{root-mean-square error}

\newacronym{ad}{AD}{automatic differentiation}
\newacronym{fd}{FD}{finite difference}

\newacronym{cfl}{CFL}{Courant--Friedrichs--Lewy}

\newacronym{pinn}{PINN}{physics-informed neural network}
\newacronym{xpinn}{XPINN}{extended \gls{pinn}}
\newacronym{lpinn}{LPINN}{Lagrangian physics--informed neural network}

\newacronym{lstm}{LSTM}{Long short--term memory}

\newacronym{rom}{ROM}{reduced order model}
\newacronym{lspg}{LSPG}{least--square Petrov--Galerkin}
\newacronym{npm}{NPM}{Neural Particle Method}
\newacronym{rbf}{RBF}{radial basis function}

\newacronym{ado}{ADO}{alternating direction optimization}

\newacronym{gep}{GEP}{gene expression programming}

\newacronym{tcr}{TCR}{transient climate response}
\newacronym{esm}{ESM}{Earth system model}

\title{Extreme Event Prediction with Multi-agent Reinforcement Learning-based Parametrization of Atmospheric and Oceanic Turbulence}

\author{
	Rambod Mojgani \\
	Department of Mechanical Engineering\\
	Rice University\\
	\texttt{rm99@rice.edu} \\
	\And
	Daniel Waelchli \\
	Chair of Computational Science\\
	ETH Zurich\\
	\texttt{wadaniel@ethz.ch} \\
	\And
	Yifei Guan \\
	Department of Mechanical Engineering\\
	Rice University\\
	\texttt{yg62@rice.edu} \\
	\And
	Petros Koumoutsakos \\
	School of Engineering and Applied Sciences\\
	Harvard University\\
	\texttt{petros@seas.harvard.edu} \\
	\And
	Pedram Hassanzadeh \\
	Department of Mechanical Engineering and \\
	Department of Earth, Environmental and Planetary Sciences\\
	Rice University\\
	\texttt{pedram@rice.edu} \\
}

\begin{document}

\maketitle

\begin{abstract}
\Glspl{gcm} are the main tools for understanding and predicting climate change. However, due to limited numerical resolutions, these models suffer from major structural uncertainties; e.g., they cannot resolve critical processes such as small-scale eddies in atmospheric and oceanic turbulence. Thus, such small-scale processes have to be represented as a function of the resolved scales via closures (parametrization). The accuracy of these closures is particularly important for capturing climate extremes. Traditionally, such closures are based on heuristics and simplifying assumptions about the unresolved physics. Recently, supervised-learned closures, trained offline on high-fidelity data, have been shown to outperform the classical physics-based closures. However, this approach requires a significant amount of high-fidelity  training data and can also lead to instabilities. Reinforcement learning is emerging as a potent alternative for developing such closures as it  requires only low-order statistics and leads to stable closures. In \gls{marl} computational elements serve a dual role of discretization points and learning agents. Here, we leverage \gls{marl} and fundamentals of turbulence physics to learn closures for canonical prototypes of atmospheric and oceanic turbulence. The policy is trained using only the enstrophy spectrum, which is nearly invariant and can be estimated from a few high-fidelity samples (these few samples are far from enough for supervised/offline learning). We show that these closures lead to stable low-resolution simulations that, at a fraction of the cost, can reproduce the high-fidelity simulations' statistics, including the tails of the probability density functions (PDFs). These results demonstrate the high potential of \gls{marl} for  closure modeling for \glspl{gcm}, especially in the regime of scarce data and indirect observations.  
\end{abstract}

\section{Introduction}


Predicting extreme weather and climate change effects demands simulations that account for complex interactions across nonlinear processes occurring over a wide range of spatiotemporal  scales. 
Turbulence as manifested in atmospheric and oceanic flows,  is a prominent example of such  nonlinear and multi-scale processes, and plays  a critical role in transporting and mixing momentum and heat in the climate system. While the governing equations of turbulent flows are known, \glspl{gcm}, which are the main tools for predicting climate variability, cannot resolve all the relevant scales. For example, in the atmosphere alone, these scales span  from $10^{-4}$ (and smaller) to $10^4$~km~\cite{Klein_ARFM_2010,vallis2017atmospheric}. Despite advances in our computing capabilities for climate modeling, this limitation is expected to persist for decades.

The effect of unresolved small scales, often referred to as \glspl{sgs}, cannot be ignored in a nonlinear system, and their two-way interactions with the resolved scales have to be accurately accounted for in order for the \glspl{gcm} to produce stable simulations and the right statistics (climate) and extreme events. Current \glspl{gcm} use semi-empirical and physics-based representations of \glspl{sgs} using closures~\cite{schneider2017earth}. The input to a closure function is the resolved scales and the output is the \glspl{sgs} fluxes of momentum, heat, etc. The current closures for many Earth system processes, particularly turbulent flows, fall short of accurately representing the two-way interactions, due to oversimplifications and incomplete theoretical understanding~\cite{rasp2018deep,hewitt2020resolving}. For example, a major shortcoming is that the current closures are too diffusive (dissipative) and also do not represent a real and important phenomenon called backscattering (basically, anti-diffusion)~\cite{Jansen_OM_2015}, preventing the \glspl{gcm} from capturing the extreme events~\cite{o2018using,guan2022stable}. Recently, there has been growing interest in using \gls{ml}, particularly \glspl{dnn}, to learn closures from data. There are two general approaches to \gls{ml}-based data-driven closure modeling:


\begin{figure}[!tb]
\centering
\includegraphics[trim=50pt 0 -50pt 0, clip, scale=0.23]{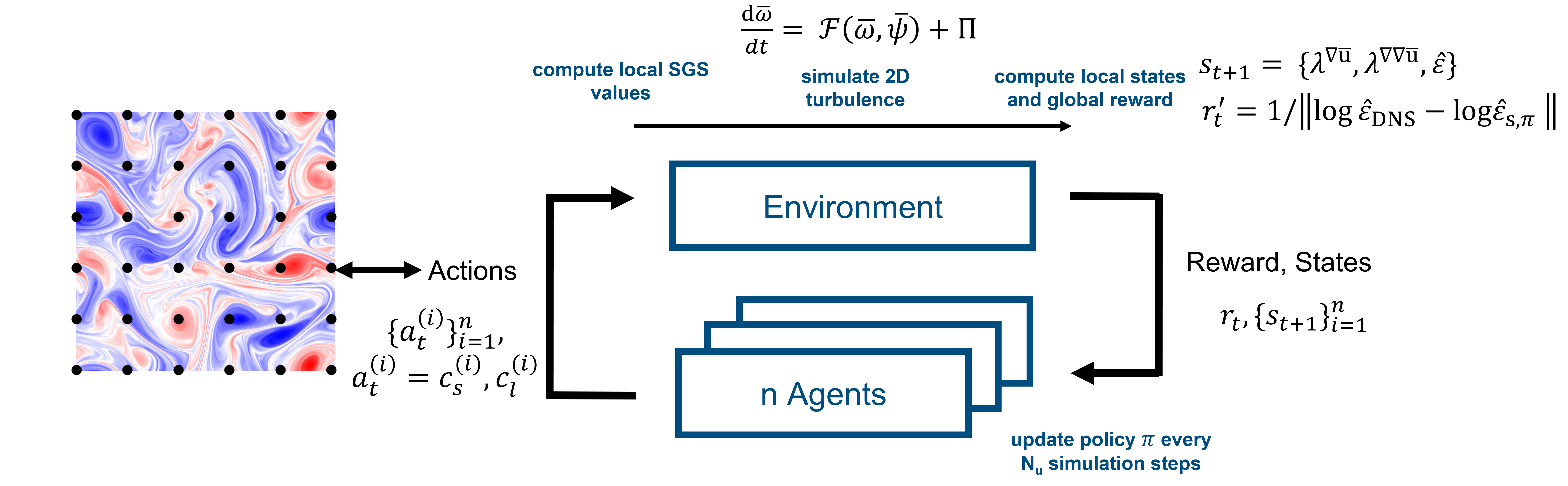}
\caption{Schematic of training an SGS closure using \gls{marl}. The invariants of gradient and Hessian of velocity ($\lambda^{\nabla \overline{u}}$ and $\lambda^{\nabla \nabla \overline{u}}$) are the states, the actions are the localized (and dynamic) coefficient $c_s(x,y,t)$ of classical Smagorinsky or $c_l(x,y,t)$ of Leith closures of the SGS term $\Pi(x,y,t)$, and the policy, $\pi$ , learns matching the enstrophy spectrum $\hat{\epsilon}$ to that of the DNS.
}
\label{fig:schematic}
\end{figure}

{\bf Supervised (offline) learning of closures:}
In this approach, many snapshots of high-fidelity data (e.g., from \gls{dns}) are collected as the ``truth'', then filtered and coarse-grained to extract the \gls{sgs} terms~\cite{grooms2021diffusion}. In turn, these data are used to train a \gls{dnn} to match the SGS terms $\Pi$ from the closures and from the truth (e.g., using a mean-square-error loss~\cite{rasp2018deep}). Once trained, the \gls{dnn} is coupled to the low-resolution solver to perform \gls{les}. Studies using a variety of architectures and test cases, from canonical turbulent systems to atmospheric and oceanic flows, have shown the possibility of outperforming classical physics-based closures such as Smagorinsky and Leith~\cite[e.g.,][]{rasp2018deep, guan2022stable, maulik2019subgrid, bolton2019applications, yuval2020stable, Srinivasan_arxiv_2023, subel2022explaining}. However, for many critical climate processes, such high-quality datasets are scarce, extracting the \gls{sgs} terms is not straightforward, and offline-learned closures can lead to unstable runs \cite{sun2023quantifying,Schneider_NCC_2023}. While adding physics to the \gls{dnn}, transfer learning, and other techniques can address these issues to some degree \cite{subel2022explaining, Guan_PhysicaD_2023, beucler2021enforcing, pedersen2023reliable, pahlavan2023explainable}, overall, offline learning remains a promising but challenging approach to data-driven SGS modeling for climate applications.   


{\bf Online learning of closures:}
Online learning is emerging as a potent alternative to supervised learning with the closures learned while they  operate on the LES.  The goal is not to match detailed flow quantities (such as the velocity field or the  \gls{sgs} terms) but instead match the low-order statistics of the high-fidelity simulations or observations. In the context of climate modeling, depending on the application, these statistics could be key properties of climate variability \cite{pahlavan2023explainable} or spectra of turbulent flows~\cite{schneider2021learning, schneider2017earth}. Online learning requires running the numerical model (e.g., a \gls{gcm}) during the DNN training, which can be challenging. In general, 3 approaches to online learning exists: 1) using a differentiable LES solver/\gls{gcm} \cite{Frezat_JAMES_2022,Shankar_arxiv_2023,Gelbrecht_EGU_2022}, 2) using \gls{eki}~\cite{kovachki2019ensemble, dunbar2020calibration, schneider2020imposing, pahlavan2023explainable}, and 3) using reinforcement learning~\cite{Novati_NMI_2021, Kurz_IJHFF_2023, Bae_NC_2022, kim2022deep}. While 1-2 have shown promising results, they face major challenges. For example, current \glspl{gcm} are not differentiable and this approach requires major development in the climate modeling infrastructure \cite{Schneider_NCC_2023}. 

Multi-agent reinforcement learning (MARL), however, has accomplished previously unattainable solutions in \gls{ml} tasks~\cite[e.g.,][]{Silver2017,Brown2019,OpenAI2019}, as well as success in improving the parametric and structural uncertainties of closures for 3D homogeneous and wall-bounded turbulent flows~\cite{Novati_NMI_2021, Kurz_IJHFF_2023, kim2022deep, Bae_NC_2022}.
However, its potential in climate-relevant applications, particularly in capturing extreme events, has remained unexplored. Here,

\begin{itemize}[noitemsep,topsep=0pt,parsep=0pt,partopsep=0pt,leftmargin=10pt]
\item We train a \gls{marl}-based \gls{sgs} closure using low-order statistics in climate-relevant flows, i.e., 2D quasi-geostrophic turbulence with different forcing or $\beta$ effects, producing multi-scale jets and vortices like those observed in the Earth's atmosphere and ocean,  
\item We remark that we use  as input states to the \gls{marl}  invariants of the flow and learn the flow-dependent coefficient $c$ of two classic physics-based closures (Leith and Smagorinsky) by matching the enstrophy spectrum the \gls{les} solver with that of the \gls{dns} (obtained from only 10 true snapshots). 
\item To test the performance of the data-driven closure, we compare the kinetic energy spectrum and vorticity PDF of the \gls{dns} with $160$ to $10240\times$ spatio-temporally coarser \gls{les}-MARL. We particularly focus on comparing the tails of these PDFs, as they represent extreme (weather) events in these prototypes. As baselines, we use localized dynamics Smagorinsky and Leith closures, which approximate $c$ as a function of the flow based on physical arguments.
\end{itemize}

\section{Scientific Multi-Agent Reinforcement Learning (SMARL)}
In deep-\gls{marl}, a \gls{dnn} is trained to learn a {\it policy} that maps the {\it states} to {\it actions}. States are fed into a \gls{dnn} and actions of the agents maximize a {\it reward}, see the schematic in~\Cref{fig:schematic}. Below, we describe the main elements of the training, for which we have utilized Korali, a general-purpose, high-performance framework for deep-\gls{marl}~\cite{Martin_CMAME_2022}.

\paragraph{\bf State:} The state vector consists of a combination of local and global variables.
As local states, instantaneous invariants $\lambda$ of filtered velocity gradients~\cite{Ling_JFM_2016} and velocity Hessians~\cite{Novati_NMI_2021} (5 non-zero local variables) are used. This choice embeds Galilean invariance into the closure. As global states, enstrophy spectrum $\hat{\epsilon}$ is used. We have found the use of these physically motivated invariants, rather than $(\bar{u},\bar{v})$ or $\bar{\psi}$ or their derivatives, to be key in learning successful closures.

\paragraph{\bf  Action:} The \gls{sgs} values at each grid point are required to evolve the governing equations of the environment in time. To retain some degree of interpretability and reduce the computational complexity of training, two classical physics-based closures are employed as the main structure of the \gls{sgs} closure: (i) Smagorinsky~\cite{Smagorinsky_MWR_1963}, which uses
$	\nu_e =   c_s \Delta^2 |\bar{\mathcal{S}} | $,
and (ii) Leith~\cite{Leith_PhysicaD_1996}, which uses
$\nu_e = c_l \Delta^3 {|\nabla \bar{\omega}|}$,
where $\nu_e(x,y,t)$ is the eddy viscosity, $\Delta$ is the filter size, $c_s(x,y,t)$ and $c_l(x,y,t)$  are the key coefficients, and 
$|\bar{\mathcal{S}} | $, and $|\nabla \bar{\omega}|$ are the magnitudes of the filtered strain rate and $\bar{\omega}$ gradient tensors. The coefficients $c$, which cannot be obtained from first principles, are considered as actions (learned as a function of the state). The actions are interpolated between the agents on to the \gls{les} grid via a bilinear scheme, and used to calculate the \gls{sgs} stress tensor, ${\tau}^{\text{SGS}} = -2\nu_e \bar{\mathcal{S}}$, which is then used to compute $\Pi$ that is needed in the low-resolution LES solver, Eq.~\eqref{eq:sys}.
\paragraph{\bf  Reward:} The goal of our \gls{sgs} closure is to match a target enstrophy spectrum, which can be calculated from a short-time high-fidelity simulation or few observations. We have computed the spectrum using {10} snapshots from a short \gls{dns} run, which is known to be insufficient to learn a successful closer offline in these flows \cite{Guan_PhysicaD_2023}. 
The reward $r_t$ at each time step is defined as the cumulative sum $r_t=\Sigma_{s=0}^{t} r'_s$ of the inverse of the $\normL{2}$ errors of the logarithms of the spectra, i.e., 
$r'_s = 1/\| \log( \hat{\varepsilon}_\text{DNS})-\log(\hat{\varepsilon}_{s,\text{RL}}) \|^2_2$, 
where $\hat{\varepsilon}_\text{DNS}$ is the time-averaged enstrophy spectrum and $\hat{\varepsilon}_{s,\text{RL}}$ is the instantaneous enstrophy spectrum at step $s$.
Note that both local variables of the state and the actions are defined at the location of the agents, and agents are uniformly distribution in the domain. 
\paragraph{\bf Environment:} The vorticity–streamfunction formulation of the 2D Navier-Stokes equation (NSE) is solved using a pseudo-spectral method. In all cases, $\Rey=20000$ and the DNS resolution is $1024$ collocation points in each direction.  The solver is coupled with Korali as the environment. Briefly, the environment provides the dynamics of \gls{les} given the actions and the states,
\begin{eqnarray}\label{eq:sys}
    \frac{\partial \overline{\omega}}{\partial t} = \mathscr{F}(\overline{\omega},\overline{\psi}) +  \Pi,
\end{eqnarray}
where $\Pi(\overline{\psi},\overline{\omega})=\nabla\times(\nabla \cdot \tau^{\text{SGS}})$, $\mathscr{F}(.)$ represents the linear and nonlinear terms of the NSE (see Eq.~\eqref{eq:NS}), and $\overline{\psi}$ and $\overline{\omega}$ are the resolved streamfunction and vorticity on the coarse grid. 

\section{Experiments}
\label{sec:expeiments}

\begin{figure}[!tbh]
\def\thirdwidth{0.32\columnwidth}
\def\myscale{0.95}
\def\myscalec{0.33}
\def\mycase{1}
\def\bottomcut{-20pt}
\begin{subfigure}[b]{0.6\columnwidth}
\includegraphics[trim=-20pt 132pt 0 -5pt, clip, scale=1.2]{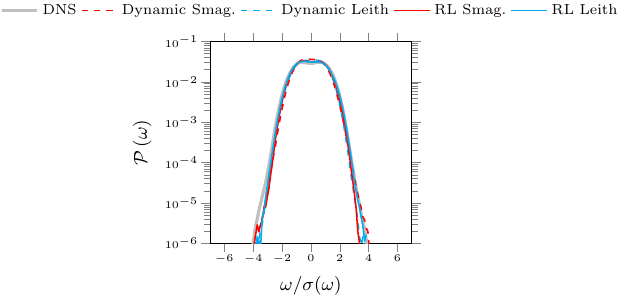}
\end{subfigure}
\begin{subfigure}[b]{\thirdwidth}
\end{subfigure}
\begin{subfigure}[b]{\thirdwidth}
\end{subfigure}	
\\
\hspace{0pt}
\vspace{-8pt}%
\begin{subfigure}[b]{0.8\columnwidth}
\caption{ Case 1: $\kappa_f=4$, $\beta=0$, $N=32$ ($10240\times$ spatio-temporally coarser)}
\end{subfigure}
\begin{subfigure}[b]{\thirdwidth}
\end{subfigure}
\begin{subfigure}[b]{\thirdwidth}
\end{subfigure}
\vspace{-8pt}
\\
\begin{subfigure}[b]{\thirdwidth}
	\centering
	\includegraphics[trim=0 -10pt 0 0, clip, scale=\myscale]{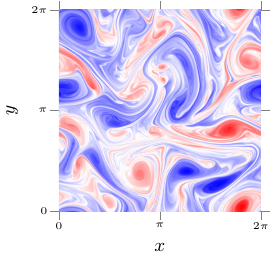}
	\label{fig:case\mycase_ens_N32}
\end{subfigure}
\hfil
\begin{subfigure}[b]{\thirdwidth}
	\centering
	\caption*{}
	\includegraphics[trim=0 0 60pt 3pt, clip, scale=\myscale]{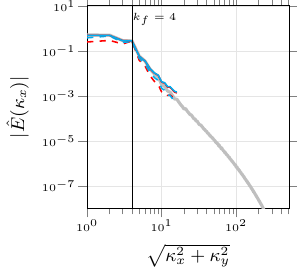}
	\label{fig:case\mycase_tke_N32}
\end{subfigure}
\hfill
\quad
\begin{subfigure}[b]{\thirdwidth}
	\centering
	\caption*{}
	\includegraphics[trim=0 0 80pt 3pt, clip, scale=\myscale]{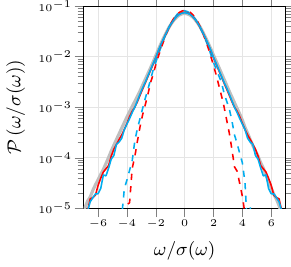}
	\label{fig:case\mycase_N32}
\end{subfigure}
\\
\def\mycase{2}
\hspace{-100pt}
\vspace{-16pt}
\begin{subfigure}[b]{0.8\columnwidth}
\caption{ Case 2: $\kappa_f=4$, $\beta=20$, $N=32$ ($10240\times$ spatio-temporally coarser)}
\end{subfigure}
\begin{subfigure}[b]{\thirdwidth}
\end{subfigure}
\begin{subfigure}[b]{0.3\textwidth}
\end{subfigure}
\\
\begin{subfigure}[b]{\thirdwidth}
\centering
	\includegraphics[trim=0 -10pt 0 0, clip, scale=\myscale]{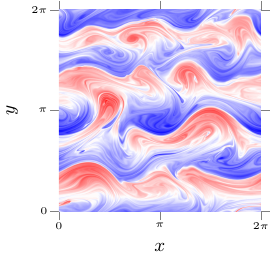}
\label{fig:case\mycase_ens_N32}
\end{subfigure}
\hfil
\begin{subfigure}[b]{\thirdwidth}
	\centering
	\caption*{}
	\includegraphics[trim=0 0 60pt 0, clip, scale=\myscale]{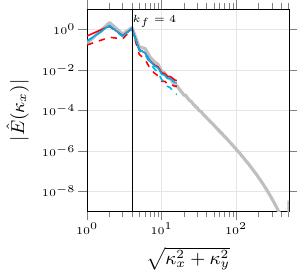}
	\label{fig:case\mycase_tke_N32}
\end{subfigure}
\hfill
\quad
\begin{subfigure}[b]{\thirdwidth}
	\centering
	\caption*{}
	\includegraphics[trim=0 0 80pt 0, clip, scale=\myscale]{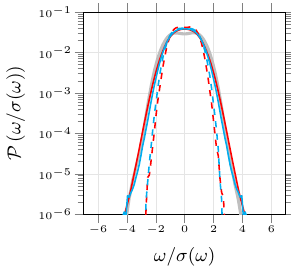}
	\label{fig:case\mycase_N32}
\end{subfigure}
\\
\def\mycase{3}
\hspace{-100pt}
\vspace{-16pt}
\begin{subfigure}[b]{0.8\columnwidth}
\caption{ Case 3: $\kappa_f=25$, $\beta=0$, $N=256$ ($160\times$ spatio-temporally coarser)}
\end{subfigure}
\begin{subfigure}[b]{\thirdwidth}
\end{subfigure}
\begin{subfigure}[b]{0.3\textwidth}
\end{subfigure}
\\
\begin{subfigure}[b]{\thirdwidth}
	\centering
	\includegraphics[trim=0 -10pt 0 0, clip, scale=\myscale]{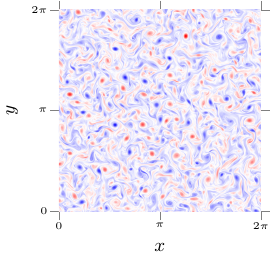}
	\label{fig:case\mycase_ens_N32}
\end{subfigure}
\hfil
\begin{subfigure}[b]{\thirdwidth}
	\centering
	\caption*{}
	\includegraphics[trim=0 0 60pt 0, clip, scale=\myscale]{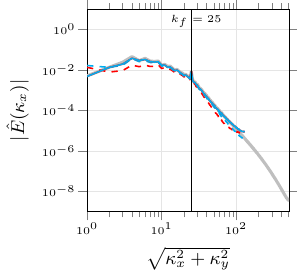}
	\label{fig:case\mycase_tke_N32}
\end{subfigure}
\hfill
\quad
\begin{subfigure}[b]{\thirdwidth}
	\centering
	\caption*{}
	\includegraphics[trim=0 0 80pt 0, clip, scale=\myscale]{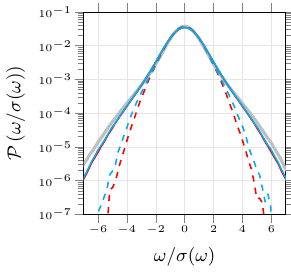}
	\label{fig:case\mycase_N32}
\end{subfigure}
\vspace{-25pt}
	\caption{
	Comparison of LES statistics with \gls{marl}-based closures and the classical closures.
	Rows (a)-(c) correspond to the cases in \Cref{tab:cases}.
}
\label{fig:cases_selected}
\end{figure}

We have developed closures for 3 different forced 2D turbulent flows on the $\beta$-plane (details in Table~\ref{tab:cases}). These cases are commonly used to evaluate the \gls{sgs} closures of geophysical turbulence~\cite[e.g.,][]{frezat2021physical, Guan_PhysicaD_2023, Srinivasan_arxiv_2023} and exhibit distinct behaviors and dynamics, as seen in snapshots of vorticity fields, $\omega$, in \Cref{fig:cases_selected}.
Training is performed with the objective of  achieving an \gls{les} enstrophy spectrum close to the target (true, DNS) spectrum. We have used the kinetic energy spectra as one unseen test metric. More importantly, the PDFs of the resolved vorticity are compared. The tails of these PDFs represent rare, extreme events, i.e., significantly large $\omega$ with a small probability of occurrence. Note that these vortices resemble the weather system's high- and low-pressure anomalies, which can cause various extreme weather events~\cite{woollings2018blocking}. LES with $16$ to $1024\times$ coarser spatial resolution and $10\times$ larger time steps coupled to the learned closures are then ran, and their statistics are compared with those of the DNS and LES with classical dynamic Smagorinsky and Leith closures. 
As summarized in \Cref{fig:cases_selected},  the tails of the vorticity PDFs clearly show the advantage of the \gls{marl}-based closures, suggesting that these closures have the right amount of diffusion and backscattering (anti-diffusion). The classical closures are too diffusive (a known problem), leading to much less frequent extreme events. The energy spectra also show the better ability of LES with  \gls{marl}-based closures in capturing the energy across the scales.

\section{Conclusion and future work}
\label{sec:conclusion}
We have trained a \gls{dnn}-based \gls{marl} to develop closures for climate-relevant turbulent flows. We show that these closures enable LES with much fewer degrees of freedom than DNS to produce statistics, including energy spectra, PDF, and most importantly, tails of the PDFs, that closely match those of the DNS. Particularly, in terms of capturing extreme events, LES with MARL-based closures significantly outperform LES with classical physics-based closures. The classical closures and even many offline-learned \glspl{dnn} produce unstable LES unless they are made overly diffusive (e.g., by eliminating backscattering), which comes at the cost of under-representing extreme events~\cite{maulik2019subgrid, guan2022stable}. With a small number of samples from DNS, which were not enough to even train an \gls{dnn} offline \cite{Guan_PhysicaD_2023}, \gls{marl} develops closures capable of capturing the statistics of such extreme events, suggesting that both diffusion and backscattering, i.e., the two-way interaction of the resolved scales and SGS, are accurately represented (analysis is in progress to quantify the interscale energy/enstrophy transfers). 

Immediate next steps include further analysis and interpretability of the $c$ distributions learned using \gls{marl} and examining the out-of-distribution generalizability of these closures. While offline-learned closures are not expected to extrapolate (e.g., to a different climate) unless methods like transfer learning are used~\cite{subel2021data}, the online-learned closures can be made generalizable by proper scaling of the invariants and spectra \cite{novati2021automating}, which might be possible with enough theoretical understanding of the changing system, e.g., the warming climate \cite{beucler2021climate}. Future work will focus on applying this framework to intermediate-complexity and comprehensive climate models to learn SGS closures (specially for offline-online learning \cite{pahlavan2023explainable}) and systematically calibrate \glspl{gcm} \cite{balaji2022general}.



\begin{ack}
PH acknowledges support from an ONR Young Investigator
Award (No. N00014-20-1-2722), a grant from the NSF CSSI Program (no. OAC-2005123),
and by the generosity of Eric and Wendy Schmidt by recommendation of the Schmidt Futures
program. PK gratefully acknowledges support from the Air Force Office of Scientific Research (MURI grant no. FA9550-21-1-005). Computational resources were provided by NSF XSEDE/ACCESS (Allocations ATM170020 and PHY220125).
\end{ack}

\medskip
\small
\setlength{\bibsep}{5pt plus 0.5ex}
\bibliographystyle{unsrtnat}
\bibliography{library.bib}{}

\appendix
\glsresetall

\section{2D turbulence}
\label{sec:turbulence}

We consider the dimensionless governing equations in the vorticity ($\omega$) and streamfunction ($\psi$) formulation in a doubly periodic square domain with length $L=2\pi$, i.e.,
\begin{subequations}\label{eq:NS}
	\begin{eqnarray}
		\frac{\partial \omega}{\partial t} + \mathcal{N}(\omega,\psi)&=&\frac{1}{\Rey}\nabla^2\omega - f -r\omega + \beta\frac{\partial \psi}{\partial x}, \label{eq:NS1}\\
		\nabla^2\psi &=& -\omega,
		 \label{eq:NS2}
	\end{eqnarray}
\end{subequations}
where
$ \displaystyle
\mathcal{N}(\omega,\psi)=
\left(\nicefrac{\partial \psi}{\partial y}\right) \left(\nicefrac{\partial \omega}{\partial x}\right) - \left(\nicefrac{\partial \psi}{\partial x}\right) \left(\nicefrac{\partial \omega}{\partial y}\right),
$
is the nonlinear advection term, 
and 
$
f(x,y) = \kappa_f[\cos{(\kappa_fx)} + \cos{(\kappa_fy)}]
$
is a deterministic forcing~\cite[e.g.,][]{chandler2013invariant,kochkov2021machine}.


To derive the equations for \gls{les}, we apply sharp spectral filtering~\cite{pope2001turbulent,sagaut2006large}, denoted by $\overline{(\cdot)}$, to Eq.~\eqref{eq:NS} to obtain
\begin{subequations}\label{eq:FNS}
	\begin{eqnarray}
		\frac{\partial \overline{\omega}}{\partial t} + \mathcal{N}(\overline{\omega},\overline{\psi})&=&\frac{1}{\Rey}\nabla^2\overline{\omega}-\overline{f}-r\overline{\omega}+\beta\frac{\partial \overline{\psi}}{\partial x}+\underbrace{\mathcal{N}(\overline{\omega},\overline{\psi}) - \overline{\mathcal{N}({\omega},{\psi})}}_{\Pi=\nabla\times(\nabla \cdot \tau^{\text{SGS}})}\label{eq:FNS1},\\
		\nabla^2\overline{\psi} &=& -\overline{\omega}\label{eq:FNS2}.
	\end{eqnarray}
\end{subequations}
The \gls{les} is solved on a coarse resolution with the \gls{sgs} term, $\Pi$, being the unclosed term, requiring a model connecting it to the resolved flow variables, i.e., closure.


For eddy viscosity models:\\
\begin{eqnarray}\label{tau}
	{\tau}^{\text{SGS}} = -2\nu_e \bar{\mathcal{S}},
\end{eqnarray}
where $\mathcal{S}_{ij}=\frac{1}{2} \left( \frac{\partial u_i }{\partial x_j} + \frac{\partial u_j }{\partial x_i}  \right)$ and $\bar{\mathcal{S}}=\sqrt{2\mathcal{S}_{ij}\mathcal{S}_{ij}}$.

\section{Test Cases}

The studied cases are summarized in Table~\ref{tab:cases}.

\def\deltatrl{\ensuremath{\Delta t_\text{RL}}}
\def\deltatdns{\ensuremath{\Delta t_\text{DNS}}}
\begin{table}[!tbh]
	\caption{The test cases and hyper-parameters in training. $\sigma$ is standard deviation.}
	\label{tab:cases}
	\centering
	\begin{tabular}{c|ccccc|cccc}
		\toprule
		Case & $\Rey$ & $\beta$ & $\kappa_f$ & $r$ &$\sigma(\omega)$ & $\deltatrl/\deltatdns$&  Training horizon & Updates policy every\\
		\midrule
		1 & $20\times10^3$ & $0$  & $4$ & $0.1$ &$5.51$&$10$&$1\times10^4\deltatrl$& $10\deltatrl$\\
		2 & $20\times10^3$ & $20$ & $4$ & $0.1$ &$10.75$&$10$&$2\times10^4\deltatrl$& $20\deltatrl$\\
		3 & $20\times10^3$ & $0$  & $25$ & $0.1$ &$13.01$&$10$&$1\times10^4\deltatrl$& $10\deltatrl$\\
		\bottomrule
	\end{tabular}
\end{table}

\end{document}